\documentclass[final]{cvpr}

\usepackage{times}
\usepackage{epsfig}
\usepackage{graphicx}
\usepackage{amsmath}
\usepackage{amssymb}
\usepackage{algorithm}
\usepackage{algpseudocode}
\usepackage[accsupp]{axessibility}  

\usepackage{hyperref}

\def\cvprPaperID{18} 
\def\confYear{CVPR 2022}

\begin{document}

\title{Epistemic Uncertainty-Weighted Loss for Visual Bias Mitigation}

\author{Rebecca S Stone, Nishant Ravikumar, Andrew J Bulpitt, David C Hogg \\
University of Leeds \\
{\tt\small \{r.s.stone, n.ravikumar, a.j.bulpitt, d.c.hogg\}@leeds.ac.uk}
}

\maketitle
\thispagestyle{empty}
\pagestyle{empty}
\begin{abstract}
Deep neural networks are highly susceptible to learning biases in visual data. While various methods have been proposed to mitigate such bias, the majority require explicit knowledge of the biases present in the training data in order to mitigate. We argue the relevance of exploring methods which are completely ignorant of the presence of any bias, but are capable of identifying and mitigating them. Furthermore, we propose using Bayesian neural networks with an epistemic uncertainty-weighted loss function to dynamically identify potential bias in individual training samples and to weight them during training. We find a positive correlation between samples subject to bias and higher predictive uncertainties. Finally, we show the method has potential to mitigate visual bias on a bias benchmark dataset and on a real-world face detection problem, and we consider the merits and weaknesses of our approach.
\end{abstract}

\section{Introduction}

Modern computer vision models are highly susceptible to learning bias and discrimination present in datasets, leading to an unfair model. While in an ideal world, an unbiased data generation or collection process would fully mitigate biases in model performance, data bias is complex and difficult to fully identify for societal and historical reasons. Even with an ideal data sampling procedure, bias can still be present.

Practitioners within the artificial intelligence community are becoming increasingly aware of gender and racial biases learned and amplified by models \cite{buolamwini2018gender, karkkainen2019fairface, georgopoulos2020investigating, bellamy2018ai, larrazabal2020gender}. Yet given the complexity and multitude of features present in visual data, there are additional biases which we may not even be aware of. Wang et al.~\cite{wang2020towards} show that simple transformations such as converting images to grayscale, taking a centre crop, and reducing image resolution all affect model fairness when introduced as a bias in the training data. In the medical imaging domain, bias has been shown to be introduced through visual artifacts~\cite{bissoto2020debiasing, bissoto2019constructing} and demographics (often a proxy for economic status)~\cite{de2019does, seyyed2020chexclusion, stacke2020measuring}. Thus, de-biasing methods which can be effective without any prior knowledge of the types and sources of bias present in the data are valuable for a variety of applications.

Modern visual systems are trained with large image datasets, where each dataset is a collection of visual attributes. A fair model is one where an input, regardless of its combination of attributes, has an equal likelihood of being assigned a correct outcome as any other input with a different set of attributes. While numerous types of bias exist (we recommend \cite{mehrabi2021survey} for a detailed discussion), we choose to focus on bias as a whole via two broad categories as follows:

\begin{enumerate}
    \item \textit{Minority group bias}. When a subgroup of the data has a particular attribute or combination of attributes which are relatively uncommon compared to the rest of the dataset, they form a minority group. A model is less likely to correctly predict for samples from a minority group than for those of the majority.
    \item \textit{Sensitive attribute bias}. A sensitive attribute (also referred to as ``protected") is one which should not be used by the model to perform the target task, but which provides an unwanted ``shortcut" which is easily learned, and results in an unfair model.
\end{enumerate}

In a scenario of complete blindness, the model is only aware of the target task, and has no prior knowledge of bias in the training set.

Most current state-of-the-art bias mitigation techniques directly use bias-informing metadata to either adjust the data before training, the model during training, or the predictions at inference time. Oversampling techniques aim to balance out the minority samples in the labelled data either before or during training to create the appearance of having a balanced set \cite{kamiran2012data}. The \textit{fairness through blindness} approach forces the model to simultaneously learn the target task and to ignore, or ``learn to not learn" a protected variable \cite{alvi2018turning}. However, these methods are susceptible to redundant encoding, where combinations of other non-protected variables act as a proxy for the protected variable. In contrast, \textit{fairness through awareness} \cite{dwork2012fairness} approaches encode and explicitly mitigate all protected features. A model is taught all subgroup and target class combinations and at inference time these relationships are removed. A recent approach titled \textit{domain independent training} aims to remove class-bias correlations by averaging class decision boundaries \cite{wang2020towards}.

Other approaches manipulate the sample feature representations in latent space to disentangle the attributes relevant for the target task from spuriously correlated features \cite{thong2021feature, tartaglione2021end}. \cite{wang2020mitigating} use a skewness-aware reinforcement learning method to determine the skewness between races on a face dataset with an adaptive margin loss function. All of these approaches \cite{thong2021feature, tartaglione2021end, wang2020mitigating} also require bias-informed metadata. 

Conversely, Amini et al.~\cite{amini2019uncovering} propose a novel method based on a variational auto-encoder (VAE) to learn latent structure simultaneously alongside the target task, and mitigate bias during training using dynamic batch selection to favor samples which are likely to be subject to bias based on their location in latent space. Recently, Xu et al.\cite{xu2021consistent} use a false positive rate penalty loss which also requires no prior knowledge of bias in the training data, and changes the objective function to reward learning a more fair model. 

We explore leveraging predictive uncertainty estimates to mitigate both sensitive attribute and minority group visual bias without any prior knowledge of the types or sources of bias in the data. Unlike the VAE used in~\cite{amini2019uncovering} which is constrained to uni-modal latent representations, we use a fully Bayesian neural network for a multi-modal posterior and better uncertainty estimates. We demonstrate on a visual bias benchmark dataset how sample-level epistemic weighting can mitigate sources of bias blindly, and further show on a real world face classification dataset how our approach can identify and mitigate sources of bias that other bias-informed methods cannot. We conclude by discussing the merits and weaknesses of our proposed approach.

\section{Related Work}
\thispagestyle{empty}
\subsection{Bayesian Neural Networks}

While Bayesian approximation via dropout \cite{gal2016dropout} has made Bayesian deep neural networks applicable to many domains due to ease of use and scalability, Markov chain Monte Carlo (MCMC) algorithms \cite{brooks2011handbook} are widely considered the gold standard for Bayesian inference. However, they are computationally intractable for large vision datasets or high-dimensional data frequently encountered in real-world computer vision applications. A well-known scalable variant of MCMC is stochastic gradient MCMC (SG-MCMC), based on diffusion processes such as the Langevin diffusion. Diffusion processes are discrete-time approximations of continuous-time processes, formulated as stochastic differential equations (SDEs) to describe the time evolution of a moving object subject to both random and non-random forces.

Given model parameters \(\theta\), dataset \(D\), prior \(p(\theta)\), and potential energy \(U(\theta)\), the posterior distribution is \(p(\theta \mid D) \propto exp(-U(\theta)) = - \textup{log}\,p(D\mid \theta) - \textup{log} \,p(\theta)\). As computing \(U(\theta)\) is not feasible for all \(D\), SG-MCMC methods approximate \(U(\theta)\) via mini-batch learning.

Welling and Teh \cite{welling2011bayesian} propose Stochastic Gradient Langevin Dynamics (SGLD) and substitute the gradient of the log-posterior density with the stochastic gradient over the minibatch and an additive Gaussian noise term that acts as an upper bound on the error. 

\begin{equation}
    \theta_i = \theta_{i-1} - \alpha_i \Delta \tilde{U}(\theta_i) + \sqrt{2 \alpha_i \epsilon_i}
\label{eq:potential}
\end{equation}

The update to parameters is shown in Equation~\ref{eq:potential}, at iteration \(i\) of the algorithm, for normal distribution \(\epsilon_i\), stepsize \(\alpha_i\) and minibatch approximation of the potential energy \(\tilde{U}\).

While convergence is in practice slower than for other MCMC algorithms, the parameter update process closely resembles stochastic gradient descent and is generalisable to any neural network. To speed up convergence and better explore complex multimodal distributions common for deep neural networks, Zhang et al. \cite{zhang2019cyclical} propose cyclical SG-MCMC (cSG-MCMC), where a cyclical stepsize schedule allows for quicker discovery of new modes. For each cycle of the learning rate schedule, an initial larger step size allows for exploration, and the subsequent smaller step sizes allow for sampling.

\subsection{Leveraging Bayesian Uncertainties}

Uncertainties can be divided into two categories, epistemic and aleatoric. Epistemic uncertainty, or model uncertainty, arises from lack of knowledge either about a process or parameter, and is caused by missing information or data. This uncertainty can be reduced given more data. Aleatoric uncertainty, or data uncertainty, arises from probabilistic variations or noise in the data and can be subdivided into two types, homoscedastic and heteroscedastic. While the heteroscedastic component varies with data, the homscedastic component, even given an infinite amount of data, cannot be reduced. \cite{ghandeharioun2019characterizing} demonstrate that epistemic uncertainty can reveal data bias, while \cite{ali2021accounting} argue that fairness approaches should equalize only errors arising from epistemic, not aleatory uncertainties. Thus, for mitigating bias, we are most interested in epistemic uncertainties; or, more generally, for the data-dependent predictive uncertainties.

Branchaud-Charron et al.~\cite{branchaud2021can} explore whether using Bayesian Active Learning by Disagreement (BALD)~\cite{gal2017deep} can help mitigate bias against a protected class. They demonstrate that an acquisition scheme which greedily reduces epistemic uncertainty has potential for bias mitigation. This method shows promising results for leveraging epistemic uncertainties to mitigate bias when all of a target class is a minority group, but does not deal with sensitive feature scenarios. Khan et al.~\cite{khan2019striking} use Bayesian uncertainty estimates to deal with class imbalance by weighting the loss function to move learned class boundaries away from more uncertain classes. While their primary focus is class uncertainties, they also consider sample-based uncertainties. They propose a curriculum learning schedule with a variational dropout Bayesian neural network which approximates uncertainties first using softmax outputs, then class-level uncertainties, and finally for the last 10 epochs, sample-based uncertainties. As we are focusing on both sensitive attribute and minority group bias, neither of which we expect to be constrained to a unique class, our approach focuses on sample-level uncertainties. This eliminates the need for tuning a curriculum learning schedule, and for encouraging the model to look at class uncertainties. Furthermore, we opt to use SG-MCMC over variational dropout for better uncertainty estimates.

\section{Methodology}
\thispagestyle{empty}
We propose a simple uncertainty-weighted loss function for visual bias mitigation. Given a Bayesian neural network with parameters \(\theta\) and posterior \(p(\theta \mid D, x) \) for class-labelled training data \(D\) and test sample \(x_i\), the predictive posterior distribution for a given predicted class \(y_i\) is then:

\begin{equation}
    p(y_i \mid D, x_i) = \int_{}^{} p(y_i \mid \theta) p(\theta \mid D, x_i) d\theta
\label{eq:posterior}
\end{equation}

We can approximate this predictive posterior via Monte Carlo sampling. Given \(T\) Monte Carlo samples total, \(\frac{T}{c}\) per learning schedule cycle \(c\), we thus have predictive mean:

\begin{equation}
    \mu_i \approx \frac{1}{T} \sum_{j=1}^{T} p(y_i \mid x_i, \theta_j)
    \label{eq:predictive_mean}
\end{equation}

and predictive uncertainty corresponding to this prediction:

\begin{equation}
    \sigma_i \approx \sqrt{ \frac{1}{T} \left(\sum_{j=1}^{T} p(y_i \mid x_i, \theta_j) - \mu_i \right)^2}
\label{eq:sigma}
\end{equation}

We propose the following uncertainty-weighted loss function for training sample \((x_i, y_i)\), given the cross entropy loss \(L(x_i, y_i)\), the additive Gaussian noise term from \ref{eq:potential} and a tunable parameter \(\kappa\) controlling the degree of weighting, especially for high-uncertainty samples.

\begin{equation}
    \hat{L}(x_i, y_i) = L(x_i, y_i)*(1.0 + \sigma_{i, y_i})^\kappa 
\label{eq:weighted_loss}
\end{equation}

As we expect a normally weighted sample to have weight 1.0, we shift the distribution such that lowest uncertainty samples are never irrelevant to the loss term. We compute \(\hat{L}\) sample-wise and then reduce over the minibatch. \(\kappa = 1\) is equivalent to a normal weighting, whereas \(\kappa \to \infty\) increases the importance of high-uncertainty samples. In our fully bias-unaware setup, \(\kappa\) is optimised based on optimal validation loss tuned via grid search.

\begin{algorithm}
\caption{Training loop using uncertainty-weighted loss}\label{alg:training}
\begin{algorithmic}[1]
\Require Training data {X, Y}, weighting parameter $\kappa$
\State Initialize parameters $\theta$
\For{each cycle, $c$}
\For{each epoch in cycle, $e$}
    \If{$e$ in sampling phase}
        \State Sample $\theta_j \sim P(\theta \mid D, x)$
        \State Save $P(y = \hat{y} \mid x, \theta_j) \, \forall x \in X$
        \State Update $\left[ w \gets w - \epsilon \nabla  L(x, y) \right]_{w \in {\theta}}$
    \ElsIf{$c$ $>$ 0}
        \State $\sigma_i \gets \sqrt{ \frac{1}{T} \left(\sum_{j=1}^{T} p(y_i \mid x_i, \theta_j) - \mu_i \right)^2}$
        \State $\hat{L}(x_i, y_i) \gets L(x_i, y_i)*(1.0 + \sigma_{i, y_i})^\kappa$
        \State Update $\left[ w \gets w - \epsilon \nabla  \hat{L}(x, y) \right]_{w \in {\theta}}$
    \EndIf
\EndFor
\EndFor
\end{algorithmic}
\end{algorithm}

The earliest moment at which we can compute \(\sigma\) over the training data is during the sampling phase of the first cycle. Uncertainty values are updated at each consecutive cycle to reflect the developing posterior, requiring a total of \(T(C - 1)\) samples from the posterior for total number of cycles \(C\). We speed up this process during training by saving each training sample's predictive distribution for each \(\theta_j\) rather than all \(\theta\). Once the sample-wise model uncertainties have been computed, all predictions can be discarded. We set the length of the sampling phase to be 5 epochs, the shortest length for which uncertainty estimates are consistently stable for a fixed seed.

\begin{table*}[th]
\label{tab:cifars_results}
\begin{center}
\begin{tabular}{llllll}
\hline
\textbf{Model} & \textbf{Description} & \textbf{Bias ($\downarrow$)} & \textbf{Mean acc (\(\%\),$\uparrow$)} & \textbf{Opp. (\(\%\),$\downarrow$)} & \textbf{Odds. (\(\%\),$\downarrow$)} \\
\hline \hline
Baseline & N-way softmax & 0.074 & 88.5 ± 0.3 & 13.07 $\pm$ 0.4 & 7.19 $\pm$ 0.2 \\ \hline
\textsc{S-Sampling} & N-way softmax & 0.066 & 89.1 $\pm$ 0.4 & 12.58 $\pm$ 0.2 & 6.91 $\pm$ 0.1 \\ \hline
\textsc{Adversarial} & w/ uniform confusion & 0.101 &  83.8 $\pm$ 1.1 & 16.71 $\pm$ 1.4 & 9.28 $\pm$ 0.7 \\
            & w/ gradient reversal, proj & 0.094 & 84.1 $\pm$ 1.0 & 14.13 $\pm$ 1.4 & 7.89 $\pm$ 0.8 \\ \hline
\textsc{DomainDiscrim} & joint ND-way softmax & 0.040 & 90.3 $\pm$ 0.5 & 7.27 $\pm$ 0.3 & 4.02 $\pm$ 0.2 \\ \hline
\textsc{DomainIndepend \cite{wang2020towards}} & joint ND-way softmax & 0.004 & 92.9 $\pm$ 0.1 & 1.07 $\pm$ 0.2 & 0.59 $\pm$ 0.1 \\ \hline
\textsc{FeatureLabel \cite{thong2021feature}} & N-way cos softmax per D & 0.004 & 91.5 $\pm$ 0.2 & 0.83 $\pm$ 0.1 & 0.46 $\pm$ 0.1 \\ \hline
\textsc{Db-Vae \cite{amini2019uncovering}} & latent structure re-weighting & 0.167 & 90.2 $\pm$ 0.4 & 6.87 $\pm$ 0.5 & 0.78 $\pm$ 0.2 \\ \hline
\textit{Our approach} & w/ unc.-weighted loss & 0.037 & 89.1 $\pm$ 0.2 & 12.12 $\pm$ 0.2 & 6.26 $\pm$ 0.2 \\ \hline

\end{tabular}
\end{center}
\caption{Multi-class classification, mean bias accuracy, equality of opportunity and equalized odds for bias benchmark dataset CIFAR-10S, a dataset with sensitive attribute bias. Note that all methods except for the baseline (a regular deterministic network with no bias mitigation), the DB-VAE, and ours are bias-informed during training.}

\end{table*}
\thispagestyle{empty}

\begin{table}[]
\label{tab:cifarm_results}
\centering
\begin{tabular}{llll}
\hline
 & \textup{Baseline} & \textit{Our approach} \\
\hline
TPR Color ($\%$),$\uparrow$) & 92.1 $\pm$ 0.1 & 93.8 $\pm$0.1 \\
TPR Gray ($\%$),$\uparrow$) & 91.6 $\pm$ 0.1 & 93.3 $\pm$ 0.1 \\
TPR Gap ($\%$),$\downarrow$) & 1.8 & 0.5 \\
\hline \\
\end{tabular}

\caption{ Uncertainty-weighted loss reduces the TPR gap on minority bias dataset CIFAR-10M.}
\end{table}

\begin{table}
\label{tab:ablation}
\centering
\begin{tabular}{llll}
\hline
 & \textup{Bias (\(\%\),$\downarrow$)} & \textup{Mean acc (\(\%\),$\uparrow$)} \\
\hline
Baseline & 0.074 & 88.5 $\pm$ 0.3  \\
cSG-MCMC & 0.060 & 88.1 $\pm$ 0.2  \\
cSG-MCMC weighted loss & 0.037 & 89.1 $\pm$ 0.2 \\
\hline \\
\end{tabular}

\caption{Ablation study showing results of a Bayesian cSG-MCMC network with regular unweighted cross-entropy loss.}
\end{table}

\section{Evaluation}

\subsection{CIFAR-10 Skewed Visual Bias Benchmark}
\thispagestyle{empty}
We compose the same baseline dataset, ``CIFAR-10 Skewed" (CIFAR-10S) proposed by Wang et al. \cite{wang2020towards} to evaluate the performance of a Bayesian CNN without any adjustments. 95\% of the images for 5 out of 10 target classes and 5\% of the remaining classes are converted to grayscale, resulting in an overall balanced dataset with respect to colour but a strong skew within each class. As a model can learn the presence or absence of colour as a class indicator, this is an instance of sensitive attribute bias. We use the official CIFAR-10 test set and a 5:1 training-validation split for the total of 60000 images.

To provide a fair comparison, we follow the same training and architecture choices as proposed, with the exception of the learning rate, which is set to maximise performance for the cyclical step size schedule of our Bayesian formulation, and with momentum 0.9 due to the Langevin dynamic estimation. Using validation loss to choose hyperparameters, we train for 280 epochs and four cycles. With the exception of our reproduction of the DB-VAE method in \cite{amini2019uncovering}, all experiments use a ResNet-18 convolutional neural network as the base architecture. For the DB-VAE, we follow the same architecture as used in the original paper, adapted for multi-target classification on CIFAR-10S, up-sizing the images to 64 x 64 to match the expected input dimensions, and grid search to find the optimal value for de-biasing parameter, $\alpha$ = 0.001.

\begin{figure*}[th]%
\begin{center}
\includegraphics[width=0.9\linewidth]{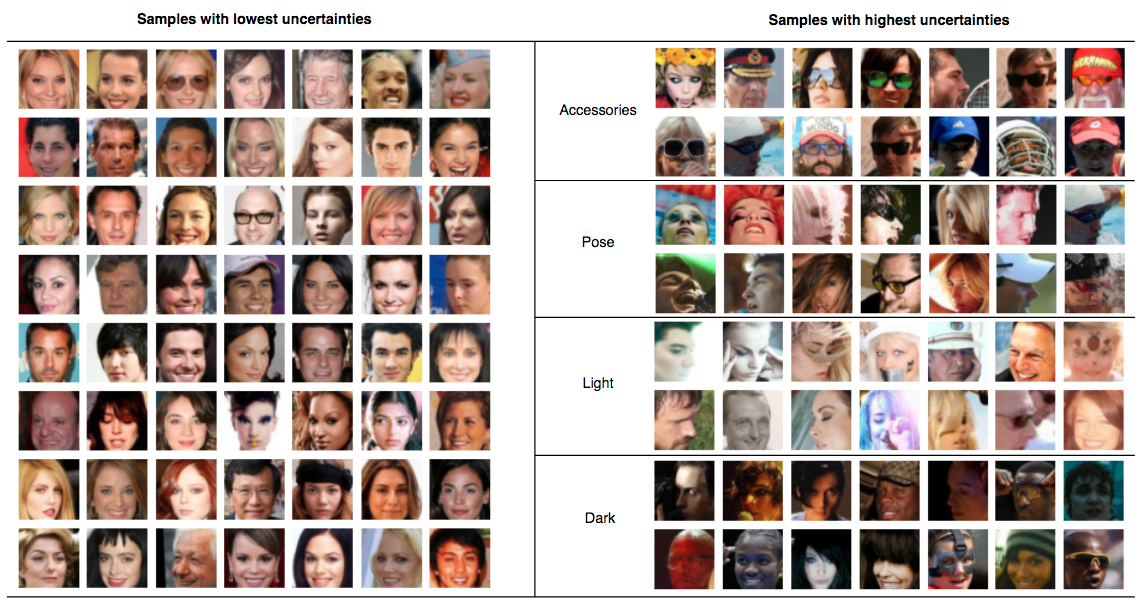}
\end{center}
\caption{Face training samples from CelebA with lowest predictive uncertainties (left) and faces with highest predictive uncertainties (right). The faces with low uncertainties tend to be well-lit, facing forward with hair cleanly framing the face, and primarily lighter-skinned with few obscuring accessories. Faces with high uncertainties are more likely to be subject to discrimination due to variance in lighting, pose, colouring, and obscuring objects, among other reasons.}
\label{fig:high_epi_faces}%
\end{figure*}

\begin{figure}%
\begin{center}
\includegraphics[width=0.45\linewidth]{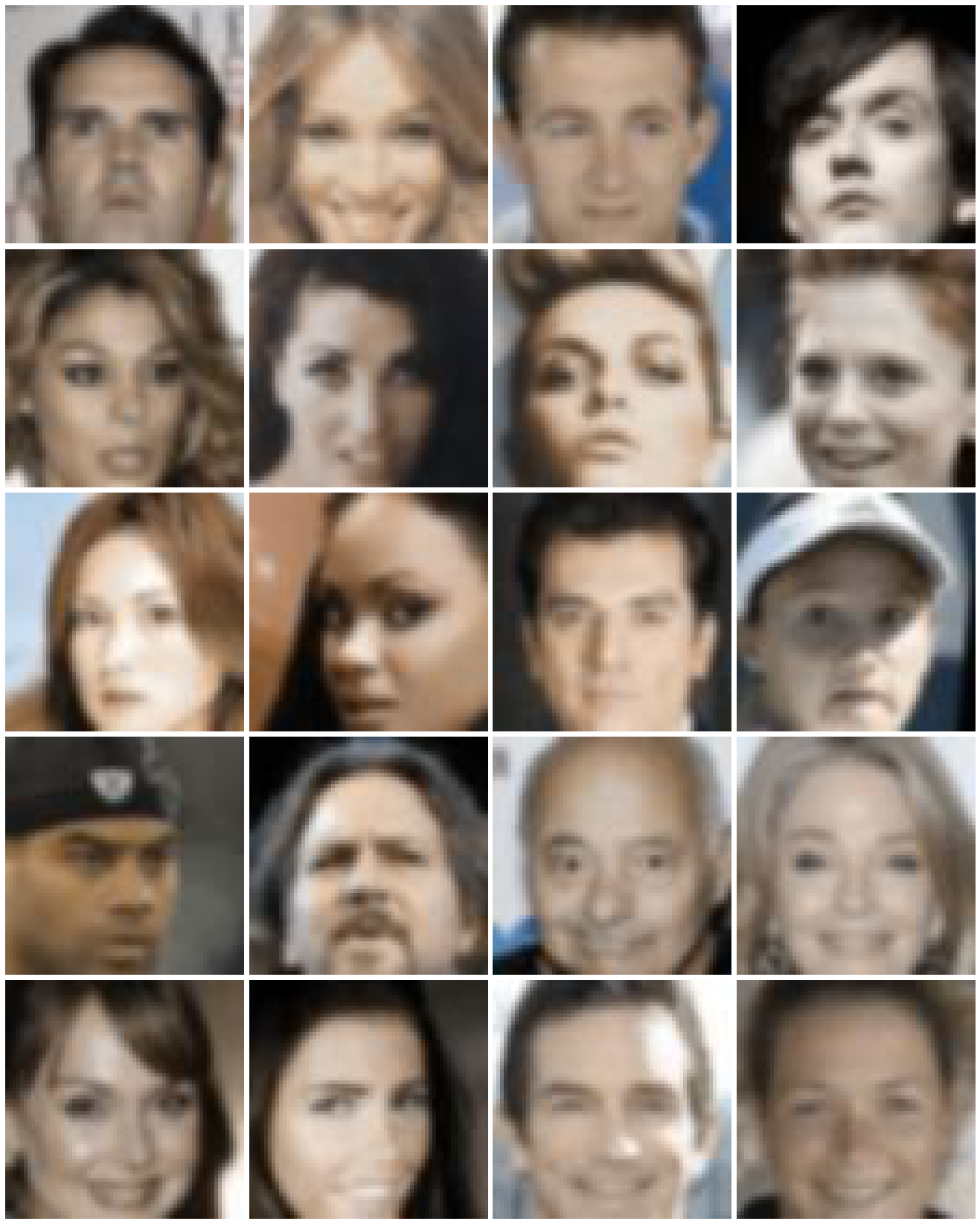}
\includegraphics[width=0.45\linewidth]{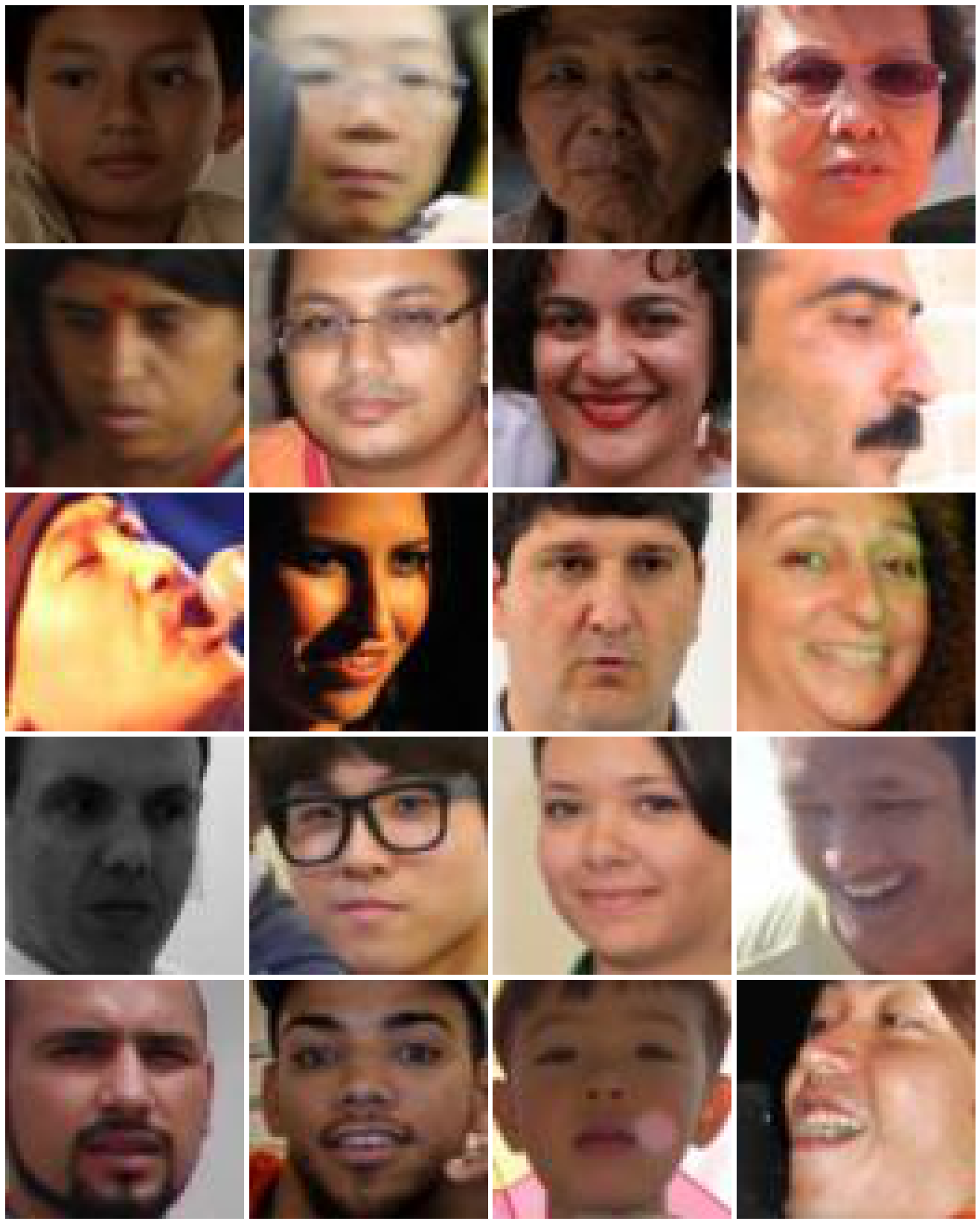}
\qquad
\end{center}
\caption{\textbf{Left:} a random selection of samples from CelebA; \textbf{Right:} and the same from FairFaces.}
\label{fig:face_datasets}%
\end{figure}

We evaluate the models using four metrics: 
\begin{itemize}
    \item \textbf{Mean test accuracy} on fully gray-scale and fully colour test sets, indicating how much the model learned the undesired correlation between the colour of each sample and its label;
    \item \textbf{Bias amplification score}~\cite{zhao2017men} proposed first in a natural language processing setting and generalised for the CIFAR-10S dataset in Equation \ref{eq:bias_amp} as suggested in \cite{wang2020towards}. \(Gr_c\) is the number of gray-scale test samples predicted as class \(c\), and \(Col_c\) the colour samples predicted as class \(c\).
    \begin{align}
    \label{eq:bias_amp}
    \textup{bias ampl.} = \frac{1}{\mid C \mid} \; \sum_{c \in C} \frac{\textup{max}(Gr_c, Col_c)}{Gr_c + Col_c} - 0.5
    \end{align}
    \item \textbf{Equalized odds} \cite{hardt2016equality} is satisfied for predictor \(\hat{y}\) when \(\hat{y}\) and sensitive attribute \textit{a} are independent conditional on outcome \textit{y}, \(\forall \gamma \in {0, 1}\): \(P(\hat{y} = 1 \mid y = \gamma, a = 0) = P(\hat{y} = 1 \mid y = \gamma, a = 1)\). A difference in equality of opportunity score is derived as per~\cite{beutel2017data} where \(FN_y^a\) is the number of false negatives of class \textit{y} with protected attribute \textit{a}:
    \begin{align}
    \label{eq:equalized_odds_score}
    \frac{1}{y} \sum_{y \in Y} \left| \frac{TP_y^1}{TP_y^1 + FN_y^1} - \frac{TP_y^0}{TP_y^0 + FN_y^0} \right|
    \end{align}
    \item \textbf{Equality of opportunity} \cite{hardt2016equality} is a relaxed form of equalized odds which requires non-discrimination on only one desired outcome, \(y = 1\), \(P(\hat{y} = 1 \mid y = 1, a = 0) = P(\hat{y} = 1 \mid y = 1, a = 1)\). A difference of equalized odds score as per~\cite{bellamy2018ai} is as follows:
    \begin{align}
    \label{eq:equality_opportunity_score}
    0.5 \ast \left( \left| FPR_y^1 - FPR_y^0 \right| + \left| TPR_y^1 - TPR_y^0 \right| \right)
    \end{align}
\end{itemize}

For CIFAR-10S, we find that samples with a sensitive attribute have higher uncertainties. They constitute 20\(\%\) of samples in the samples with the highest 10\(\%\) uncertainties, while only 5\(\%\) of samples in the dataset have a sensitive attribute. In contrast, less than 2\(\%\) of samples in the lowest 10\(\%\) uncertainties have a sensitive attribute.

As CIFAR-10S is a case of sensitive attribute bias, we formulate a second dataset CIFAR-10M ``Minority" (CIFAR-10M) to represent minority group bias. We set grayscale as the minority attribute, and for each of the 10 classes, \(5\%\) of samples are converted to grayscale. The remaining \(95\%\) remain in colour. We split the training-validation sets using a 5:1 ratio. The results of our debiasing method on this dataset are presented in Table~\ref{tab:cifarm_results}.

While not competitive with all bias-informed methods, our approach demonstrates an ability to de-bias blindly on both the benchmark dataset with sensitive attribute bias (CIFAR-10S), and our constructed dataset with minority group bias (CIFAR-10M).

\subsection{On a Real-World Face Detection Problem}
\thispagestyle{empty}
Similar to \cite{amini2019uncovering}, we create a face vs. no-face binary classification dataset using 20k instances of faces from CelebA~\cite{liu2018large} and 20k non-face samples from a variety of different classes from ImageNet~\cite{deng2009imagenet}, for a training set of 40k images. We evaluate the model on FairFaces~\cite{karkkainen2019fairface}\footnote{Due to the inaccessibility of the Pilot Parliaments Benchmark dataset~\cite{buolamwini2018gender} because of data privacy issues, we have chosen to evaluate on FairFaces.}, consisting of 108,501 images with gender, race, and age annotations, balanced across 7 race groups: White, Black, Indian, East Asian, Southeast Asian, Middle Eastern, and Latino, and 9 age subgroups from ``0-2" to ``over 70". The high level of diversity is visible in Figure~\ref{fig:face_datasets}. The distribution shift between the training and test data is such that we center crop the CelebA images to 124 x 124 and downsample all face and non-face training data to 64 x 64 such that the general resolution and positioning of the face in the images do not hinder the model from generalising. We use the Fréchet Inception Distance~\cite{heusel2017gans} between the CelebA and FairFace datasets to guide our transforms to the CelebA data, supported by visual comparison of random selections of images from both datasets. No transforms other than resizing are applied to the FairFace data, ensuring that the diversity is still intact.

We train a regular ResNet18 to convergence at 30 epochs using an 8:2 split of the CelebA/ImageNet dataset for training and validation and a standard SGD optimizer with learning rate of 0.01. The cSG-MCMC Bayesian model with uncertainty-weighted loss is similarly trained with 4 cycles of 30 epochs each for a total of 120 epochs.

\begin{figure}%
\begin{center}
\includegraphics[width=1.0\linewidth]{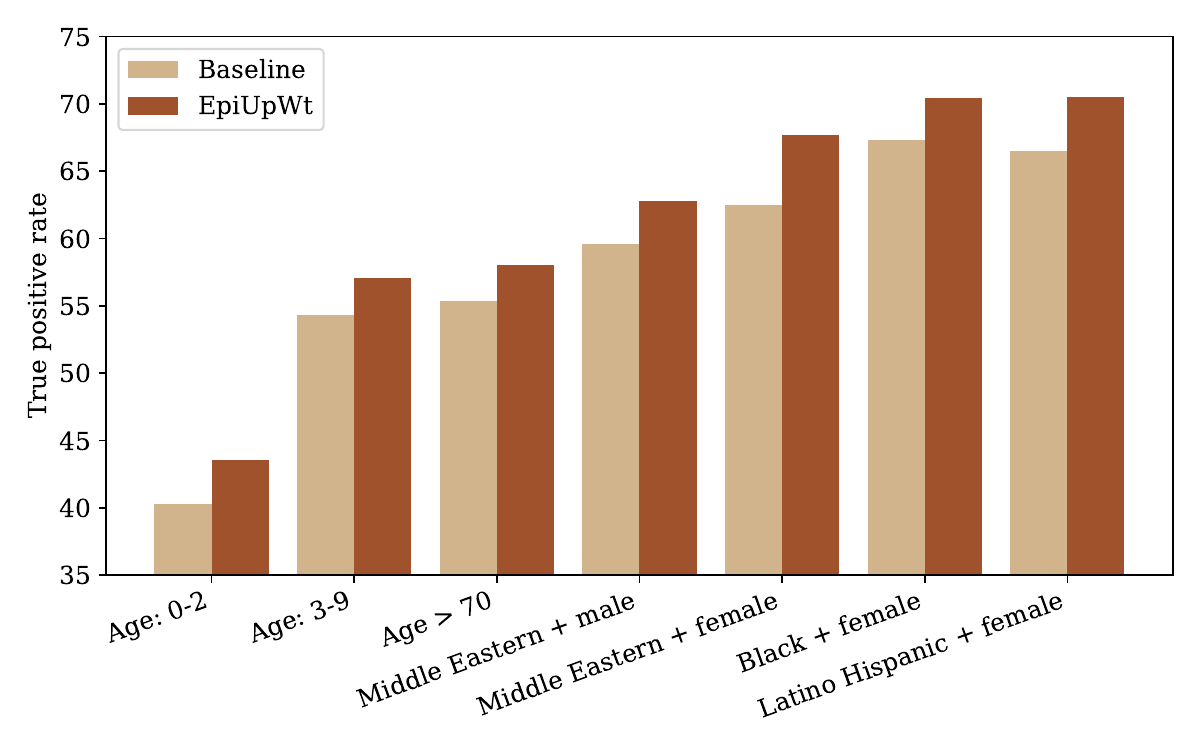}
\end{center}
\caption{Performance discrepancies between baseline (deterministic model with no de-biasing) and Bayesian model with uncertainty-weighted loss for minority subgroups with lowest TPRs.}
\label{fig:ff_subgroup_accs}%
\end{figure}

For every subgroup, the uncertainty-weighted loss decreases the TPR gap, with the discrepancies for the 7 subgroups with lowest TPR rates shown in Figure~\ref{fig:ff_subgroup_accs}.

\begin{figure}%
\begin{center}
\includegraphics[width=0.8\linewidth]{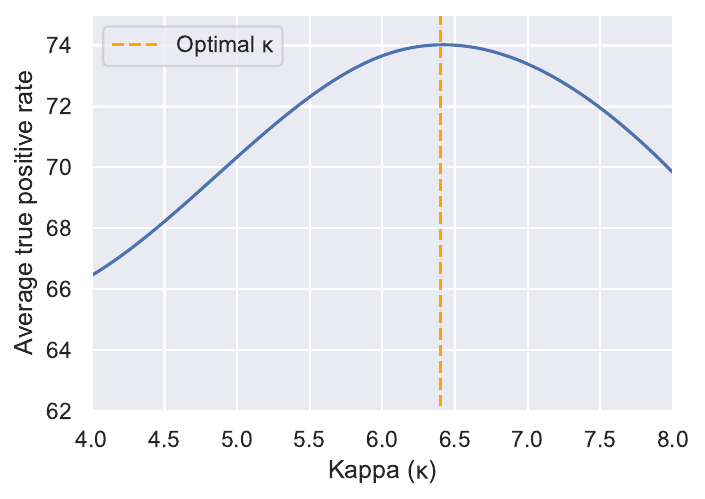}
\end{center}
\caption{True positive rate (TPR) over the entire FairFace dataset as a function of tunable de-biasing parameter kappa $\kappa$, showing how the degree of de-biasing can be controlled by $\kappa$.}
\label{fig:tuning_kappa}%
\end{figure}

Given that sample-level weighting by a factor of \(N\) during training is equivalent to that sample appearing \(N\) times, our approach could be categorized as a type of sub-sampling algorithm. Thus, it suffers from the same weakness as all sub-sampling algorithms, a tendency to overfit over-sampled data. This can only be partially mitigated by aggressive data augmentation. We hypothesize that this explains why increasing tunable de-biasing parameter $\kappa$ beyond the optimal value results in worse performance as shown in Figure~\ref{fig:tuning_kappa}.

Figure~\ref{fig:high_epi_faces} shows samples with high uncertainties, which clearly have features which make them more likely to be subject to bias. A bias-informed method could strongly mitigate bias due to known societal biases such as gender and race, or skin phenotype. But since it would be unlikely to also have access to meta-data which identifies variances in lighting, pose, image resolution, etc., all of which also result in unfairness, such methods would not target such biases.

Such an approach is valuable in medical imaging applications with large population image analysis due to the inherent difficulty in collecting meta data. Sensitivity and privacy requirements result in imaging datasets with very few annotations and little, if any, associated patient meta data. This presents a challenge for bias-informed methods, and serves as motivation for further exploration of methods which can mitigate without requiring comprehensive knowledge of all biases.

\section{Conclusion}
\thispagestyle{empty}
We have shown that a predictive uncertainty-weighted loss function has potential for bias mitigation for datasets with unknown sources of bias. We cannot conclude that the approach works in all cases based on our experiments, nor do we attempt to provide a mathematical proof that this should be the case. Some training datasets may contain an over-sampling from unprivileged groups, in which case the correlation with epistemic uncertainties may no longer exist. Thus, our exploration focuses only on cases of minority group and sensitive attribute bias. While outperformed by most bias-informed models, our method is a step towards exploring how predictive uncertainties in Bayesian neural networks can be leveraged for identifying, understanding, and mitigating the types and sources of visual bias in data.

\textbf{Acknowledgements.} The first author is a recipient of a Rabin Ezra Scholarship. Special thanks to Jose Sosa and Mohammed Alghamdi from the School of Computing's Computer Vision Group for critical feedback and many great discussions.

\small
\bibliographystyle{ieee_fullname}

\end{document}